\documentclass[letterpaper, 10 pt, conference]{ieeeconf}  

\IEEEoverridecommandlockouts                              

\usepackage{graphicx}
\usepackage{tabularx}
\usepackage{booktabs}
\usepackage{multirow}
\usepackage[utf8]{inputenc}
\usepackage[T1]{fontenc}
\usepackage{amsmath,amssymb}
\usepackage{booktabs}
\usepackage{geometry}
\usepackage{soul,xcolor}
\usepackage{minted}
\usepackage{booktabs}
\usepackage{siunitx}
\let\labelindent\relax
\usepackage{enumitem}
\usepackage{array}
\usepackage{cite}
\usepackage[hidelinks]{hyperref}
\usepackage{xurl} 
\usepackage{microtype} 
\usepackage{censor}
\usepackage{makecell}
\newcolumntype{C}[1]{>{\centering\arraybackslash}p{#1}}
\DeclareRobustCommand{\change}[1]{{\sethlcolor{white}\hl{#1}}} 

\title{\LARGE \bf
Programming Manufacturing Robots with Imperfect AI:\\ 
LLMs as Tuning Experts for FDM Print Configuration Selection}

\author{Ekta U. Samani$^{1}$ and Christopher G. Atkeson$^{1}$
\thanks{This work was supported by the Robotics and AI Institute.}
\thanks{$^{1}$E. U. Samani and C. G. Atkeson are with the Robotics Institute at Carnegie Mellon University,
        {\tt\small \{esamani,cga\}@andrew.cmu.edu}}%
}

\begin{document}

\maketitle
\thispagestyle{empty}
\pagestyle{empty}


\begin{abstract}
We use fused deposition modeling (FDM) 3D printing as a case study of how manufacturing robots can use imperfect AI. In FDM, print configuration strongly affects output quality. Yet, novice users typically rely on default configurations, trial-and-error, or direct recommendations from generic AI models (e.g., ChatGPT). These strategies can produce complete prints, but they do not reliably meet specific objectives. We present a modular approach that treats an LLM as a source of tuning expertise. We embed this source of expertise within a Bayesian optimization loop. An approximate evaluator scores each candidate print configuration and returns structured diagnostics, which the LLM uses to propose natural-language adjustments that are compiled into machine-actionable guidance for optimization. On 100 Thingi10k parts, our LLM-guided loop achieves the best configuration on 78\% objects with 0\% likely-to-fail cases, while direct AI model recommendations are rarely best and exhibit 15\% likely-to-fail cases. These results suggest that LLMs provide more value as constrained decision modules in optimization loops than as end-to-end oracles for print configuration selection. We expect this result to extend to broader LLM-based robot programming.
\end{abstract}

\section{Introduction}

3D printers that use Fused Deposition Modeling (FDM), also known as Fused Filament Fabrication (FFF), are special-purpose manufacturing robots that deposit molten plastic along a toolpath to rapidly turn geometric specifications into usable physical parts. For many consumer prints, default printer-filament profiles in the toolpath-generation software (slicer) are sufficient to succeed. The harder, practical problem is what happens when the goal is not merely ``it printed,'' but achieving a specific objective (e.g., quality prioritized over time and cost) across diverse geometries. Here, the gap between novices and expert workflows is stark: novices tend to rely on default profiles and, increasingly, direct recommendations from generic AI models (e.g., ChatGPT), whereas experts tune print configurations by diagnosing defects in completed prints and making targeted parameter adjustments. \cite{kwon20243dpfix}. 

This paper investigates using a large language model (LLM) as an imperfect source of expertise. Instead of relying on the AI model's internal reasoning to output a full print configuration, we use the LLM as a constrained decision-maker inside an optimization loop. Specifically, we score candidate print configurations with a toolpath-based evaluator that returns a scalar objective and structured diagnostics. The LLM uses these diagnostics to select a primary issue and propose a small set of parameter edits, which we transform into guidance that steers optimization. Across a diverse set of real-world parts, our approach improves consistency relative to baselines reflecting common non-expert workflows such defaults and direct recommendations from AI models. Our contributions are:

\begin{itemize}[leftmargin=*]
    \item We introduce a modular, LLM-guided optimization loop for selecting FDM print configurations.
    \item We develop an approximate evaluator for FDM print configurations that computes structured diagnostics from toolpaths.
    \item We propose a guidance compiler that converts the LLM's natural-language guidance into machine-actionable guidance for Bayesian optimization.
    \item We demonstrate improved print configuration recommendations, achieving the best configuration on 78\% objects with 0\% likely-to-fail cases.
\end{itemize}

\section{Related Work}

\subsection{FDM Parameter Tuning and Orientation Selection}
Extensive research shows that FDM outcomes depend on print parameters (e.g., layer height, infill, thermal/cooling conditions), affecting geometry, surface properties and mechanical performance \cite{mengesha2022comprehensive}. In addition to print parameters, part orientation during printing is also a first-order decision variable, balancing surface finish (staircase), support requirements, and build time \cite{thrimurthulu2004optimum, buj2019influence}. Models link staircase roughness to layer thickness and local surface slope, motivating orientation- and layer-height-aware planning \cite{pandey2003improvement}. Recent work optimizes orientation (sometimes jointly with other parameters), typically by fitting empirical models to a small set of measured responses (e.g., roughness, build time) on limited test geometries \cite{hooshmand2021optimization}. FDM parameter settings reported as optimal are not universal and often do not transfer across printing setups or objects \cite{golab2022generalisable}. We therefore optimize orientation and print parameters separately for each object.

\begin{figure*}[ht]
    \centering
    \includegraphics[width=\textwidth]{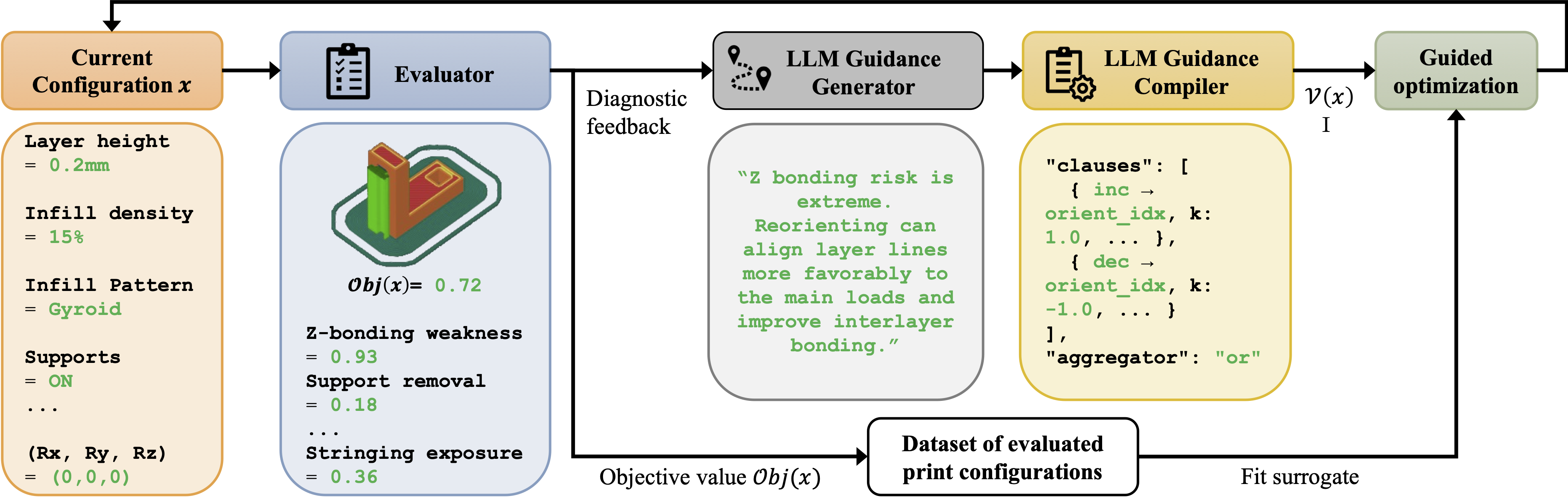}
    \caption{Overview of the modular LLM-guided optimization loop for FDM 3D print configuration selection. An evaluator computes objective value $Obj(x)$ and structured diagnostics, an LLM proposes corrective changes, and a compiler maps these changes to guidance ($\mathcal{V}(x), \mathcal{I}$) used for optimization. Lighter boxes denote example intermediate outputs.}
    \label{pipeline}
\end{figure*}

\subsection{Printability, Quality Metrics, and Failure Prediction}
Prior work evaluates print feasibility and quality using geometry-derived features, toolpath reasoning, and sensing. Toolpath-aware analyses show that the realized deposition plan dominates defects and failure mechanisms \cite{lensgraf2016beyond}. Data-driven models predict surface roughness from process variables and geometric features \cite{wu2019predictive}. Monitoring-and-control systems use in-process imaging to detect and mitigate failures \cite{brion2022generalisable}. High-fidelity thermo-mechanical simulations are informative \cite{digimat}, but their computational cost motivates using faster surrogates for iterative search. Fast evaluators that provide feedback for iterative correction in diverse parts remain rare \cite{he2025development}. We address this gap with a manually-designed evaluator that vetoes configurations that are likely to fail and returns proxy risk penalties computed from an approximation of the layerwise deposition. We also use the same manually-designed evaluator for quantitative assessment, we are not aware of a widely accepted metric, and human evaluation and physical testing are too expensive.

\subsection{LLMs in 3D Printing and Optimization}
LLMs are being explored in 3D printing, spanning benchmark/task collections \cite{eslaminia2024fdm}, G-code comprehension \cite{jignasu2023towards}, and monitoring/control agents that generate corrective actions from images of the print state \cite{jadhav2024llm}. Adjacent work explores defect prediction from process parameters \cite{pak2025additivellm}. However, part-conditioned optimization of process parameters prior to printing remains underexplored. In other domains, recent work studies LLMs as iterative optimizers conditioned on prior trials \cite{yang2023large}, tools that augment Bayesian optimization \cite{liu2024large,agarwal2025searching,chang2025llinbo,wang2024large}, and hybrid frameworks that decouple prior reasoning from acquisition using updated posteriors \cite{gupta2025llms}. We apply this principle to FDM print configuration selection by restricting the LLM to provide guidance while optimization is primarly driven by an updated posterior.

\section{Method}

Given part geometry and user objectives (e.g., prioritizing quality over time and cost), we cast print configuration selection as an optimization problem. In each iteration, a candidate configuration is evaluated to produce a scalar objective and diagnostics on printability and quality issues. An LLM prioritizes which issue to address next and proposes parameter-level changes. These proposals are converted into guidance for generating the next configuration in two forms: soft guidance that biases generation toward configurations consistent with the proposed changes and hard constraints that restrict the next step to modifying only the parameters named by the selected actions, i.e., the implicated parameters. The process repeats under this combined scheme. The following subsections describe each module, and Fig. \ref{pipeline} summarizes the overall approach.

\vspace{-1mm}
\subsection{Evaluator}
\label{sec:evaluator}

We manually created an evaluator suitable for FDM print configuration selection. Given a configuration $x$, the evaluator runs the toolpath-generation software (slicer) and computes structured diagnostics, returning (i) a scalar score for optimization, (ii) feasibility vetoes, and (iii) issue-level penalties. The optimization objective is
\begin{equation}
Obj(x)
= w_t\frac{t(x)/t_{\mathrm{r}}}{1+t(x)/t_{\mathrm{r}}}
+
w_c\frac{c(x)/c_{\mathrm{r}}}{1+c(x)/c_{\mathrm{r}}}
+
w_qQ(x),
\label{eq:objective}
\end{equation}
where $t(x)$ is the slicer-estimated print time, $c(x)$ is the estimated filament cost, $t_{\mathrm{r}}$ and $c_{\mathrm{r}}$ are reference values used for bounded normalization, and $w_t$, $w_c$, and $w_q$ are user-defined weights on print time, cost, and quality, respectively. The term $Q(x)$ is an approximate quality penalty computed from bounded diagnostic penalties $p_i(x)\in[0,1]$ grouped into surface-geometry artifacts $S$, functional performance $F$, and finish/post-processing burden $P$. We also define feasibility veto tests $v_j(x)$ based on discrete print failure modes. If any veto triggers, we declare $x$ infeasible and set $Q(x)=+\infty$. 
If no veto triggers, we summarize $p_i(x)$ within each group with a smooth maximum (log-mean-exp) and convert each group score into a clipped, normalized excess $e_k(x)\in[0,1]$ (i.e., overshoot) above a fixed goal  for $k\in\{S,F,P\}$. We aggregate these group excesses as follows,
\begin{equation}
Q(x)=\max_k e_k(x)+\frac{\lambda}{3}\Bigl(1-\max_k e_k(x)\Bigr)\sum_k e_k(x).
\label{eq:quality_scalar}
\end{equation}

\noindent We compute $p_i(x)$ and $v_j(x)$ from slicer warnings, a layerwise deposition approximation, and PLA-specific heuristics; details are in the Appendix.

\subsection{LLM Guidance Generator}
The LLM guidance generator takes as input the evaluator’s structured diagnostics and a list of admissible corrective actions, each defined as a change to a print parameter. For a print configuration $x$, let $\zeta(x)$ denote the vector of tunable print parameters. Each action modifies one entry of $\zeta(x)$, and the LLM may propose one or more actions. The list is intentionally restricted to a small set of high-leverage parameters spanning continuous, integer-valued, binary, and categorical controls. This restriction ensures that every recommendation is concrete and implementable, and that its effects are reliably reflected in the diagnostic signals available under our approximate evaluator. The guidance generator is prompted to perform a constrained decision: it identifies exactly one primary issue to address and proposes corrective actions, each expressed either as a directional adjustment (increase or decrease of a parameter) or as a categorical switch (changing a discrete option). It outputs the selected actions and their rationales for auditability, and the guidance compiler parses them.

\subsection{LLM Guidance Compiler}
\label{sec:parser}

The guidance compiler converts the LLM guidance generator's corrective actions into two artifacts: (i) a differentiable soft-violation score $\mathcal{V}(x)\in[0,1]$ that measures how strongly a candidate configuration $x$ contradicts the proposed changes to $\zeta(x)$, and (ii) an implicated-parameter set $\mathcal{I}$ (the union of parameters referenced by the selected actions), which enables hard constraints by freezing $\mathcal{I}^c$ while optimizing over $\mathcal{I}$.

For each selected action, we produce an action-level violation score $\mathcal{V}_{\text{action}}(x)\in[0,1]$ and an action-level implicated set (the parameters referenced by that action). The translation proceeds in three steps:
\begin{enumerate}[leftmargin=*]

  \item The action is rewritten into a small set of \emph{atomic} clauses (minimal edit statements) and an aggregator that specifies whether the clauses are intended conjunctively or disjunctively. Each clause identifies the target setting(s) in $\zeta(x)$ and a change type (e.g., increase/decrease, categorical switch).

  \item Each clause is matched to a residual template in Table~\ref{residualtable}, yielding a nonnegative residual $\rho_i(s(x))\ge 0$ and clause metadata (e.g., importance and confidence). We write $\rho_i(s(x))$ with the understanding that only the parameters referenced by the clause are used. The implicated set for the action is the union of the referenced parameters across its clauses.

  \item Residuals are normalized as $\tilde\rho_i=\rho_i/(1+\rho_i)\in[0,1)$, and each clause is assigned a weight $w_i>0$ from its metadata. Clause violations are then aggregated using the operators in Table~\ref{aggregatortable}.
\end{enumerate}

\begin{table}[]
\centering
\caption{Residual templates for the LLM Guidance Compiler}
\label{residualtable}
\small
\setlength{\tabcolsep}{5pt}
\begin{tabularx}{\columnwidth}{@{}>{\raggedright\arraybackslash}X >{\raggedright\arraybackslash}X c@{}}
\toprule
\makecell[l]{\textbf{Change type}\\(example cues)} & \makecell[l]{\textbf{Residual}} & \textbf{Params} \\
\midrule
\makecell[tl]{Directional preference\\(increase $u$ by $k$)} &
\makecell[tl]{\\ $\mathrm{ReLU}(k-\Delta u)$\\} & \makecell[tl]{\\ $k$} \\
\midrule
\makecell[tl]{Equality/target\\(set $u=\alpha$)\\($u$ should equal $v$)} &
\makecell[tl]{\\ $(u-\alpha)^2$\\$(u-v)^2$} & \makecell[tl]{\\ $\alpha$\\} \\
\midrule
\makecell[tl]{Range/box\\($u\in[L,U]$; $u\ge L$)} &
\makecell[tl]{$\mathrm{ReLU}(L-u)$\\$+\;\mathrm{ReLU}(u-U)$} & \makecell[tl]{$L,U$} \\
\midrule
\makecell[tl]{Margin\\(make $u\ge v+\delta$)} &
\makecell[tl]{$\mathrm{ReLU}((v+\delta)-u)$} & \makecell[tl]{$\delta$} \\
\midrule
\makecell[tl]{Ratio target ($u/v\approx\beta$)} &
\makecell[tl]{$\left(u/(v+\varepsilon)-\beta\right)^2$} & \makecell[tl]{$\beta$} \\
\midrule
\makecell[tl]{Sum cap ($\sum_i u_i \le k$)} &
\makecell[tl]{$\mathrm{ReLU}\!\left(\sum_i u_i-k\right)$} & \makecell[tl]{$k$} \\
\midrule
\makecell[tl]{Sequence monotonicity\\($a\le b\le c$)} &
\makecell[tl]{$\sum_i \mathrm{ReLU}(u_i-u_{i+1})$} & \makecell[tl]{--} \\
\bottomrule
\end{tabularx}
\end{table}

Across the selected actions, we aggregate action-level violations as $\mathcal{V}(x)=1-\prod(1-\mathcal{V}_{\text{action}}(x))$ and take $\mathcal{I}$ as the union of all parameters referenced by the actions. Together, $\mathcal{V}(x)$ and $\mathcal{I}$ provide the interface between the guidance generator output and the acquisition optimizer.

\begin{table}[]
\centering
\caption{Violation aggregators for the LLM guidance compiler}
\label{aggregatortable}
\small
\setlength{\tabcolsep}{5pt}
\begin{tabularx}{\columnwidth}{@{}>{\raggedright\arraybackslash}X >{\raggedright\arraybackslash}X c@{}}
\toprule
\makecell[l]{\textbf{Aggregator}\\(example cues)} & \makecell[l]{\textbf{Form}} & \textbf{Params} \\
\midrule
\makecell[tl]{Soft-AND (conjunction):\\``all of \ldots''} &
\makecell[tl]{$ 1 - \prod_i (1-\tilde \rho_i)^{w_i}$} &
\makecell[tl]{$w_i$} \\
\midrule
\makecell[tl]{Soft-OR (disjunction/ \\ at least one):\\``$u$ or $v$, any of these''} &
\makecell[tl]{\\ $\prod_i (\tilde \rho_i)^{w_i}$} &
\makecell[tl]{\\ $w_i$} \\
\bottomrule
\end{tabularx}
\end{table}

\subsection{Guided Optimization}
\label{sec:acquisition}

At each iteration, we fit a probabilistic surrogate model to the observations and select the next configuration by maximizing a weighted criterion. Starting from a standard base acquisition $\alpha(x)$ (e.g., expected improvement), we apply soft guidance by downweighting candidates according to the guidance compiler’s violation score, $\tilde{\alpha}(x)=\alpha(x)\exp(-\eta\,\mathcal{V}(x))$ with $\eta>0$, which biases selection toward configurations consistent with the proposed changes while still allowing deviations when the predicted gain is large. We apply hard constraints by restricting the acquisition to candidates that modify only the implicated parameters in $\mathcal{I}$ while holding all other print parameters fixed to their current values. 

\vspace{-1mm}

\section{Experiments}
\vspace{-1mm}
\subsection{Dataset}
We use objects from the Thingi10k dataset \cite{zhou2016thingi10k} for evaluation. It contains parts from an online 3D-printing community platform and represents models encountered ``in the wild” for 3D printing. We randomly select 100 single-component parts with fewer than 100 vertices and fewer than 100 faces. This choice aligns the part complexity with the fixed coarse rasterization used by our evaluator. \change{A finer rasterization would allow evaluation of parts with finer geometric detail.}

\vspace{-1mm}
\subsection{Implementation Details}
\label{sec:impl}

We score each configuration with the evaluator in Section \ref{sec:evaluator}. It includes fixed thresholds and logistic scales, which we set using three hand-designed parts to ensure that each diagnostic exhibits sensible trends under controlled perturbations. The tunable configuration $\zeta(x)$ consists of a discrete build orientation (three Euler-style rotations) and 13 print parameters: layer height, first layer height, first layer extrusion width, infill density, infill pattern, brim width, support material (binary), number of perimeters, number of bottom solid layers, number of top solid layers, filament maximum volumetric speed, elephant foot compensation, and seam placement \cite{prusaknowledgebase}. In Eq. \ref{eq:objective}, we set $w_q$, $w_t$, and $w_c$ as $0.8$, $0.1$, and $0.1$, respectively. References $t_{\mathrm{r}}$ and $c_{\mathrm{r}}$ are the medians of slicer-estimated print time and filament cost over 32 Sobol samples \cite{gpyopt} from the configuration space.

\subsubsection{Baselines}

We evaluate four baselines reflecting common user workflows: (i) default parameters (from the printer-filament profile) in the mesh's as-provided orientation, (ii) heuristic reorientation with the same defaults, and (iii-iv) direct recommendations of AI models. For all baselines, we use the slicer to generate toolpaths from the suggested configuration, apply slicer-recommended adjustments (e.g., adding a brim), and report results using the adjusted configuration.

For heuristic reorientation, we enumerate the six axis-aligned rotations mapping each mesh $\pm X$, $\pm Y$, and $\pm Z$ direction to global $+Z$ (``up''). For each candidate, we voxelize the rotated mesh and compute a support proxy: the area of downward-facing exposed voxel faces not on the lowest $z$-slice (treated as bed contact). We choose the orientation minimizing this unsupported downward area, breaking near-ties by maximizing estimated bed-contact area. All other print parameters remain at defaults. 

To emulate a novice user seeking high-quality AI guidance, we query two current-generation, high-capability AI models, ChatGPT 5.2 Thinking \cite{openai_chatgpt_tool} and Gemini 3 Pro \cite{google_gemini_tool}, via their paid chat interfaces (Plus and Edu, respectively) with reasoning enabled. For each object, given a mesh in STL format and the default printer-filament profile, the model returns exactly one print configuration for the same weighted objective as our method (favoring quality over time and cost). The prompt restricts choices to parameters in $\zeta(x)$ and explicitly permits computing basic geometric metadata.

\subsubsection{Optimization variants}
 
We run black-box optimization using GPyOpt with expected improvement as the base acquisition function and a Mat\'ern 5/2 kernel with automatic relevance determination \cite{gpyopt}. We warm-start the surrogate with 16 initial evaluations, then run 40 sequential iterations. Primary results use GPT-5.2 via the API (medium reasoning) \cite{openai_gpt52_system}, with in-context examples linking diagnostic penalties to corrective actions and up to two corrective actions per iteration. To avoid API costs, ablations use a quantized Llama 3.1 70B model \cite{grattafiori2024llama,hugging} locally on a workstation with two NVIDIA RTX 5090 GPUs. Ablations also include optimization with handcrafted guidance as a baseline: a human expert provides guidance in the same format as the LLM, compiled by the guidance compiler. \change{Our code and prompts can be found at} \url{https://ektas0330.github.io/llm_guided_optimization/}.

\vspace{-1mm}
\subsection{Results}

\vspace{-1mm}
\subsubsection{Key results}

We compare modular LLM-guided optimization to baselines that approximate non-expert workflows for selecting a print configuration. Each method outputs one configuration per object, which we evaluate using $Obj(x)$ (Eq.~\ref{eq:objective}). \change{The chat-based baselines produce a recommendation once, while the optimization-based methods use evaluator feedback over multiple iterations before returning a final recommendation.} For optimization, we run 10 random initializations and summarize the performance of each object by the median final objective across initializations. Table~\ref{motivationtable} reports the median objective across objects (lower is better), object-level best/near-best rates among the listed methods, and the fraction flagged \change{by the evaluator} as likely to fail (feasibility veto(es) triggered). 

\begin{table}[]
\centering
\caption{End-to-end performance of print-configuration selection methods, evaluated with $Obj(x)$ (Eq.~\ref{eq:objective}).}
\label{motivationtable}
\scriptsize
\renewcommand{\arraystretch}{0.9}
\setlength{\tabcolsep}{3pt}
\begin{tabularx}{\columnwidth}{@{}>{\raggedright\arraybackslash}Xccccc@{}}
\toprule
\multirow{2}{*}{Method} & \multicolumn{1}{c}{Median} & \multicolumn{3}{c}{\makecell[c]{\% of objects that are best\\or within $x\%$ of best}} & \multirow{2}{*}{\makecell[c]{\change{Flagged} \\ likely\\to fail}} \\ \cmidrule(lr){3-5}
 & \multicolumn{1}{c}{$Obj(x)$} & Best & Within 1\% & Within 5\% & \\ \midrule
\makecell[l]{Original orientation\\w/ default parameters} & 0.37 & 0  & 2  & 12 & 6  \\
\makecell[l]{Heuristic orientation\\w/ default parameters} & 0.26 & 3  & 4  & 15 & 9  \\
\makecell[l]{ChatGPT 5.2 \\Thinking}                           & 0.25 & 8  & 20 & 50 & 15 \\
\makecell[l]{Gemini 3 Pro\\(Chat Interface)}              & 0.21 & 4  & 18 & 48 & 15 \\
\midrule
\makecell[l]{(Unguided)\\Optimization}                   & 0.24 & 7  & 14 & 44 & \textbf{0} \\
\makecell[l]{\textbf{GPT-5.2 guided}\\\textbf{Optimization (Ours)}} & \textbf{0.14} & \textbf{78} & \textbf{82} & \textbf{90} & \textbf{0} \\ \bottomrule
\end{tabularx}
\end{table}

\noindent \textbf{LLM-guided optimization finds the best configuration most often with 0\% likely-to-fail cases.}
Across the 100-object dataset, our method achieves the best objective value (lowest) among the listed approaches on 78\% of objects and is within 1\% / 5\% of the lowest on 82\% / 90\%. It also attains the lowest median objective value, indicating that improvements are not limited to a small fraction of objects. It yields 0\% likely-to-fail cases under our feasibility-veto criterion.

\noindent \textbf{Default parameters and direct recommendations deom AI models are often suboptimal and sometimes likely-to-fail.} Default print parameters (original or heuristic reorientation) are within 5\% of the best on only 12\% and 15\% of objects, respectively, and show non-trivial likely-to-fail rates (6-9\%). Direct recommendations from AI models (ChatGPT 5.2 Thinking, Gemini 3 Pro) improve median objective relative to defaults and are within 5\% of the best on 50-48\% of objects, but they are rarely best and incur a 15\% likely-to-fail rate.

\begin{figure}[t]
    \centering
    \includegraphics[width=0.74\columnwidth]{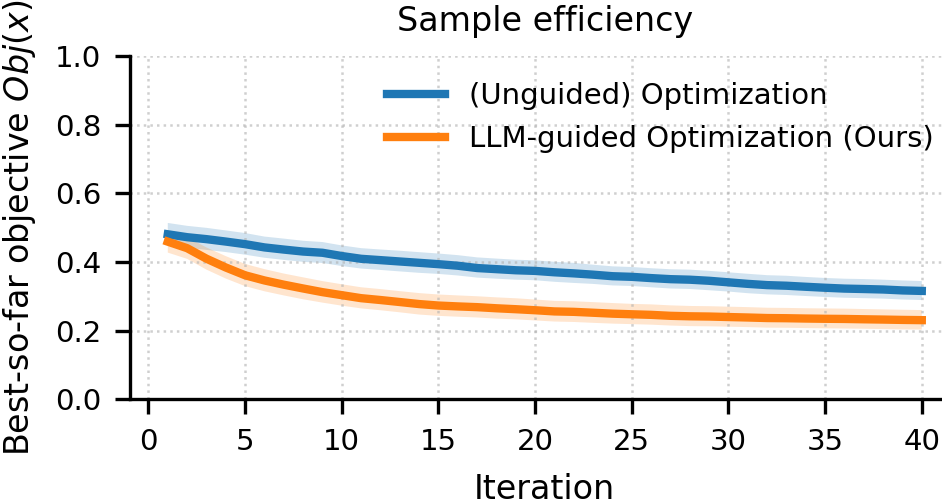}
    \caption{Sample efficiency of guided (ours) vs. unguided optimization. Curves show best-so-far $Obj(x)$ averaged over 10 initializations and objects. Shading shows 95\% bootstrap confidence intervals over objects.}
    \label{fig:sample_efficiency}
\end{figure}

\noindent \textbf{LLM guidance improves final outcomes and sample efficiency relative to unguided optimization.}
\change{Unguided optimization achieves a median objective of 0.24 and is within 1\% / 5\% of the best on 14\% / 44\%}. With LLM guidance, performance improves markedly: the median objective drops to 0.14, and the method is best on 78\% of objects and within 5\% of the best on 90\%. As shown in Fig.~\ref{fig:sample_efficiency}, LLM guidance also improves sample efficiency, reaching lower best-so-far objective values in fewer iterations than unguided optimization.

\subsubsection{Ablation studies}

We test three guidance design choices: guidance source, whether the prompt includes in-context examples, and the per-iteration action budget, in Tables~\ref{tab:winrates_source}, \ref{fewshotvszeroshot}, and \ref{tab:winrates_bandwidth}, respectively. Each table reports a pairwise win-rate matrix, computed as follows.

For each object, we compare variants pairwise across 10 random initializations. Variant $i$ wins an initialization against $j$ if it achieves a lower final objective by more than a fixed tie tolerance. Outcomes within tolerance are treated as ties and count as wins for both variants. Variant $i$ wins the object if it wins more initializations than $j$. Aggregating over objects yields a win-rate matrix $P_{ij}$ (ties counted as $0.5$), and we summarize each method by its mean $P_{ij}$ against all other methods.

\noindent \textbf{Guidance matters more than which LLM, and LLM guidance matches handcrafted guidance.}
With prompting and action budget fixed (examples, two actions), GPT-5.2 (medium reasoning) guidance, Llama 3.1 70B guidance, and handcrafted guidance beat the no-guidance variant on 92\% of the objects (Table~\ref{tab:winrates_source}). Guided variants differ modestly: GPT slightly outperforms Llama, and both are close to handcrafted guidance.
\begin{table}[]
\centering
\caption{Pairwise win-rate matrix for guidance-source ablation.}
\label{tab:winrates_source}

\setlength{\tabcolsep}{2.5pt}       
\renewcommand{\arraystretch}{1.08}  

\begin{tabular}{@{}C{0.27\columnwidth}|cccc|c@{}}
\toprule
\makecell[c]{Row beats\\ column} &
\makecell[c]{None} &
\makecell[c]{Llama 3.1\\70B} &
\makecell[c]{GPT-5.2\\API} &
\makecell[c]{Hand-\\crafted} &
\makecell[c]{Mean \\ winrate}  \\
\midrule
None                        & --     & 0.08  & 0.08  & 0.08 & 0.08 \\
\midrule
Llama 3.1 70B        & 0.92  & --     & 0.46 & 0.50 & 0.63 \\
\midrule
GPT-5.2 API & 0.92  & 0.54 & --     & 0.54 & 0.67 \\
\midrule
Handcrafted               & 0.92 & 0.50 & 0.46 & --     & 0.63 \\
\bottomrule
\end{tabular}
\end{table}

\noindent \textbf{In-context examples improve guidance quality.}
Holding the model and action budget fixed (Llama 3.1 70B, two actions), including examples of appropriate corrective actions in the prompt beats omitting them on 62\% of objects (Table~\ref{fewshotvszeroshot}). Prompting \textit{without examples} also underperforms handcrafted guidance on 61\% of objects,, indicating that a small number of examples makes the guidance signal more reliable for iterative optimization.

\begin{table}[]
\centering
\caption{Pairwise win-rate matrix for prompting ablation (with vs.\ without examples). LLM is Llama 3.1 70B.}
\label{fewshotvszeroshot}
\setlength{\tabcolsep}{2.5pt}       
\renewcommand{\arraystretch}{1.08}  

\begin{tabular}{@{}C{0.27\columnwidth}|ccc|c@{}}
\toprule
\makecell[c]{Row beats \\ column} & 
\makecell[c]{LLM prompted \\ w/o examples} & 
\makecell[c]{LLM prompted \\ with examples} &
\makecell[c]{Hand- \\crafted} &
\makecell[c]{Mean \\ winrate}  \\
\midrule
LLM prompted w/o examples  & --    & 0.38  & 0.39  & 0.38  \\
\midrule
LLM prompted with examples & 0.62  & --     & 0.50 & 0.56 \\
\midrule
Handcrafted                & 0.61 & 0.50 & --     & 0.56 \\
\bottomrule
\end{tabular}%
\end{table}

\noindent \textbf{More actions per iteration improves performance.}
With model and prompting fixed (Llama 3.1 70B with examples), performance improves as the per-iteration action budget increases (Table~\ref{tab:winrates_bandwidth}). Allowing one action yields a modest gain over zero actions. Allowing two actions is much stronger, beating the one-action setting on 75\% of the objects and nearing handcrafted guidance. We do not consider budgets above two actions because actions are restricted to directional/categorical changes to a small set of parameters, and two actions capture most issue-specific corrections in our formulation.

\begin{table}[]
\centering
\caption{Pairwise win-rate matrix for action-budget ablation (0/1/2 actions per iteration). LLM is Llama 3.1 70B.}
\label{tab:winrates_bandwidth}

\setlength{\tabcolsep}{2.5pt}       
\renewcommand{\arraystretch}{1.08}  

\begin{tabular}{@{}C{0.27\columnwidth}|cccc|c@{}}
\toprule
\makecell[c]{Row beats\\ column} &
\makecell[c]{None} &
\makecell[c]{LLM: \\ one action} &
\makecell[c]{LLM: \\ two actions} &
\makecell[c]{Hand-\\crafted} &
\makecell[c]{Mean \\ winrate}  \\
\midrule
None                        & --     & 0.32  & 0.08  & 0.08 & 0.16 \\
\midrule
LLM: one action        & 0.68  & --     & 0.24 & 0.22 & 0.38 \\
\midrule
LLM: two actions & 0.92  & 0.76 & --     & 0.50 & 0.72 \\
\midrule
Handcrafted               & 0.92 & 0.78 & 0.50 & --     & 0.74 \\
\bottomrule
\end{tabular}
\end{table}

\subsubsection{\change{Qualitative} physical validation}
\label{sec:physical_validation}
\begin{figure*}[t]
    \centering
    \includegraphics[width=0.8\textwidth]{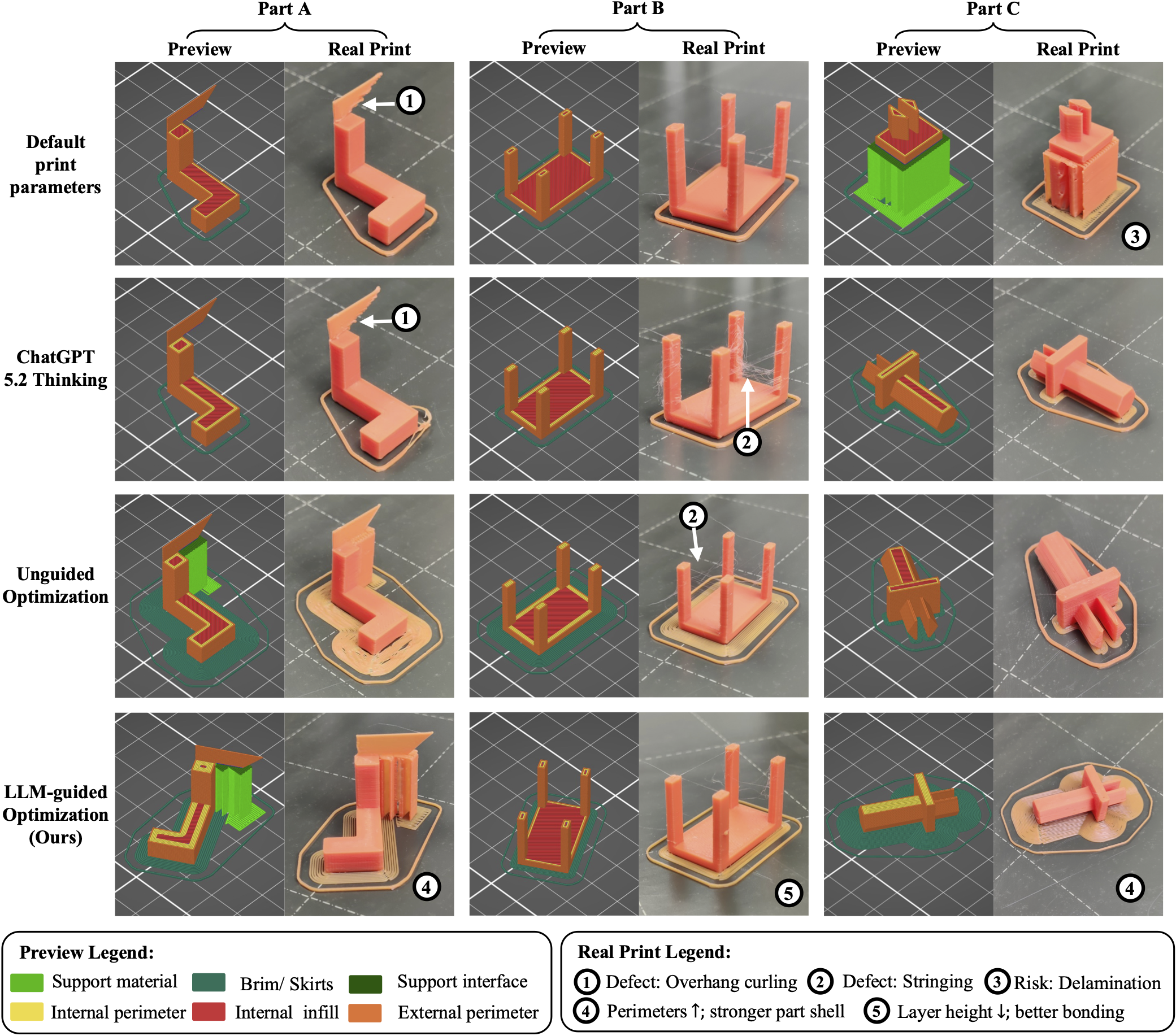}
    \caption{Qualitative comparison of toolpath previews and photos of 3D prints for three representative parts across methods. Our method uses GPT-5.2 (medium reasoning) guidance. Numbered callouts mark key print issues and configuration changes. Preview colors follow the legend.}
    \label{fig:results}
\end{figure*}

All quantitative comparisons consistently use our evaluator to assign an overall score to each print configuration since a widely accepted scoring metric is not available. For qualitative validation, we physically print a subset of objects on a Prusa i3 MK3S printer using the recommended print configuration of each method. Fig. \ref{fig:results} \change{shows how the resulting prints differ. For Part A, the default and ChatGPT configurations leave the overhang unsupported and it curls, while our method adds support to avoid it. For Part B, the ChatGPT and unguided optimization configurations produce prints with stringing between the posts, whereas our method's configuration does not led to this defect, while improving bonding by choosing a lower layer height. For Part C, the default configuration is flagged for delamination risk, while our method chooses a different orientation and increases perimeters to strengthen it.}

\vspace{-1mm}
\section{Discussion}
\vspace{-2mm}
\subsection{Design choices}
We structure the method as a modular loop that cleanly separates evaluation, guidance, and optimization. This modularity lets us swap in higher-fidelity simulators or alternative acquisition strategies without changing the LLM's role: a constrained decision-maker operating over structured diagnostics. The residual templates and soft logic aggregators in our guidance compiler, inspired by probabilistic soft logic \cite{bach2017hinge}, can also be replaced with other functional forms as needed. Although we instantiate the loop for PLA on a Prusa i3 MK3S, the method is not printer- or material-specific as it only requires an evaluator calibrated to the target process that provides reliable feasibility vetoes and bounded, issue-level diagnostics. Higher-fidelity evaluators can capture additional defect mechanisms but typically increase computation cost, so we adopt an approximate evaluator that is fast enough for iterative search while still emitting actionable signals. However, evaluation remains expensive due to repeated toolpath analysis, making evaluation efficiency a first-order concern. We therefore use Bayesian optimization, which is effective when only a small number of evaluations are feasible. Together, these choices isolate the central question of the paper: Does explicit evaluation and optimization improve engineering decision-making relative to direct recommendations based solely on the internal reasoning of an AI model? The evaluator and optimization routine used here are thus one practical instantiation of that procedure, not a claim of universal optimality. More generally, the same procedure can be used in other robot-controlled processes, by pairing an LLM that contributes imperfect but useful knowledge with an evaluator that encodes specific process knowledge from human designers.

\vspace{-1mm}
\subsection{Limitations and Scope}
A key limitation of the current work is that we collapse heterogeneous objectives and defect mechanisms into a scalar. Although each risk term is normalized to $[0,1]$, the penalties reflect qualitatively different phenomena with different functional forms, and our aggregation imposes fixed trade-offs that may not match all use cases. More broadly, parameter tuning is inherently multi-objective, and scalarization obscures the Pareto structure over quality, time, and cost. Second, we tune only a subset of print parameters: the full parameterization includes controls whose effects may not be identifiable from our evaluator's diagnostics, so some valid interventions are out of scope and the loop may miss improvements that require print parameters not represented in $\zeta(x)$ or interactions not captured by the diagnostics. Finally, experts often improve reliability via part redesign, which we do not consider in this work. Future work, outlined in the next section, addresses several of these limitations.

\section{Conclusion}
In this paper, we explored whether embedding LLM guidance within an evaluation-and-optimization loop improves FDM print configuration selection relative to direct recommendations produced using only an AI model's internal reasoning. We presented a modular approach that couples an approximate toolpath-driven evaluator, which emits structured diagnostics, with an interface that compiles constrained natural-language adjustments into machine-actionable guidance for Bayesian optimization. On 100 Thingi10k parts, our approach selected the best configuration for 78\% of objects with 0\% likely-to-fail cases, while chat-based AI model recommendations were rarely best and produced 15\% likely-to-fail cases. These results suggest that LLMs can contribute more effectively as constrained decision modules inside evidence-driven optimization loops than as end-to-end oracles for FDM print configuration selection. More broadly, these findings motivate treating imperfect AI models as components in evaluation-grounded pipelines for robots, rather than as standalone oracles. Future work will extend the guidance generator to use multimodal evidence (e.g., renderings and toolpath previews), study alternative role allocations for the LLM within the loop (including evaluator and optimization-agent roles), replace the current evaluator with higher-fidelity simulators or learned predictors, and move from scalarization to multi-objective optimization that returns a Pareto set for downstream selection.

\vspace{-1mm}
\section*{Appendix: Evaluator Details}
\label{app:evaluator_details}

\textit{Feasibility vetoes:} We veto prints flagged by slicer warnings (low bed adhesion, collapsing overhangs, long bridges) and add two computed vetoes: unsupported islands and slender tower instability. Unsupported islands are disconnected components that start without overlap with recently deposited material. We convert their area and unsupported height into a bounded risk score aggregated across islands. Slender tower instability detects tall, thin features by tracking connected regions across layers, estimating width from cross-sections, and converting height-to-width ratios into bounded risk scores that are aggregated across features. 

\textit{Surface-geometry artifacts:} We quantify surface artifacts from layer discretization with a staircasing penalty. For each mesh face with slope angle $\theta_v$ relative to the build axis, we compute a terrace-amplitude proxy $a=h|\cos\theta_v|(1-|\cos\theta_v|)$ (with layer height $h$) and take the area-weighted 95th percentile over faces (after excluding a small band near the build plate). We normalize it with a fixed reference and clamp to obtain $p_{\mathrm{stair}}\in[0,1]$.

\textit{Functional performance:} We compute four penalties that capture common ways process parameters affect mechanical integrity and dimensional fidelity: strength deficit, z-bonding weakness, perimeter-infill decoupling, and XY-dimensional risk. Strength deficit uses an effective strength index that combines a shell term (perimeters and top/bottom skins, with mild line-width/nozzle adjustments) and an infill term (infill fraction modulated by an infill-pattern factor and mild process modifiers). We map this index to a bounded shortfall relative to a fixed adequacy target and clamp to obtain $p_{\mathrm{strength}}\in[0,1]$. Z-bonding weakness targets delamination by combining three cues that typically limit inter-layer adhesion: poor contact between successive layers, insufficient time/thermal conditions for bonding, and operation near volumetric flow limits. Each cue is converted to a bounded sub-score, and the aggregate is dominated by the worst cue. 
We then apply a saturating transform, amplified for load-bearing parts whose load direction aligns with the build axis, and clamp to obtain $p_{\mathrm{zbond}}\in[0,1]$. Perimeter-infill decoupling measures how weakly the interior fill is tied into the perimeter shell. We estimate perimeter-interior contact from layerwise raster masks and normalize it by the contact that would be possible given the geometry (the contact opportunity). We convert the fraction into a penalty and clamp to obtain $p_{\mathrm{pi}}\in[0,1]$, with $p_{\mathrm{pi}}=0$ when there is no interior contact opportunity. XY-dimensional risk models first-layer-driven distortion from excess first-layer squish, $|z|$ offset, and first-layer flow, with a temperature-dependent bulging term \cite{prusaknowledgebase}. We attenuate it using capped elephant-foot compensation and early-layer concealment (area growth) \cite{protolabs}, scale by footprint sensitivity (perimeter-to-area statistics), and clamp to obtain $p_{\mathrm{xy}}\in[0,1]$.

\textit{Finish/post-processing burden:} We capture nuisance defects/cleanup difficulty with two penalties, stringing and support removal. We compute the fraction of travel motion that occurs in free space relative to total extrusion and map it with a bounded logistic to get $p_{\mathrm{stringing}}\in[0,1]$. Support removal scores the extent and structure of the support-part interface using contact density with stacking and fragmentation modifiers, mapped with saturating transforms to obtain $p_{\mathrm{support}}\in[0,1]$.

\vspace{-1.1mm}
\bibliographystyle{IEEEtran}
\bibliography{references}

\end{document}